\documentclass{article} %
\usepackage[T1]{fontenc}
\usepackage{iclr2027_conference,times}

\usepackage{amsmath,amsfonts,bm}

\def\eqref#1{equation~\ref{#1}}

\def\1{\bm{1}}

\DeclareMathAlphabet{\mathsfit}{\encodingdefault}{\sfdefault}{m}{sl}
\SetMathAlphabet{\mathsfit}{bold}{\encodingdefault}{\sfdefault}{bx}{n}

\usepackage{hyperref}
\usepackage{url}
\usepackage{booktabs}
\usepackage{graphicx}
\usepackage{placeins} %
\usepackage{array}
\usepackage{wrapfig}
\usepackage{xcolor}
\usepackage{enumitem}

\setcitestyle{numbers,square,citesep={,}}\makeatletter\def\NAT@sort{1}\def\NAT@cmprs{1}\makeatother
\providecommand{\ourwork}{RefRoute}
\providecommand{\ourdata}{\ourwork-Data}
\newcommand{\SamplingSteps}{50}

\newcommand{\RecordedScalingSteps}{50}

\newcommand{\LoRARank}{128}
\newcommand{\TrainingLR}{10^{-4}}

\newcommand{\PerGPUBatch}{1}

\newcommand{\StageOneGPUs}{4}
\newcommand{\StageTwoGPUs}{8}

\newcommand{\TargetResolution}{1024}
\newcommand{\ReferenceDownsample}{4}
\newcommand{\VAECompression}{16}

\newcommand{\NativeGridSide}{\number\numexpr\TargetResolution/\VAECompression\relax}
\newcommand{\CompactGridSide}{\number\numexpr\TargetResolution/\ReferenceDownsample/\VAECompression\relax}
\newcommand{\NativeRefTokens}{\number\numexpr\NativeGridSide*\NativeGridSide\relax}
\newcommand{\CompactRefTokens}{\number\numexpr\CompactGridSide*\CompactGridSide\relax}

\newcommand{\IllustrativeReferences}{16}
\newcommand{\IllustrativeNativeTokens}{\number\numexpr\IllustrativeReferences*\NativeRefTokens\relax}

\newcommand{\CompressorLayers}{6}
\newcommand{\CompressorKernel}{3}
\newcommand{\CompressorStride}{2}
\newcommand{\CompressorAttentionBlocks}{2}

\newcommand{\draftnote}[1]{\par\noindent{\small\color{orange!85!black}\textbf{Draft note:} #1}\par}
\newcommand{\tbd}[1]{{\color{orange!85!black}[#1]}}

\newcommand{\CheckSamplingProtocol}{%
  \ifnum\SamplingSteps=\RecordedScalingSteps\relax\else
    \draftnote{The planned protocol uses \SamplingSteps\ steps, but the saved
    scaling data use \RecordedScalingSteps\ steps. Rerun the scaling experiment
    before using its values as results for the new protocol.}%
  \fi}

\providecommand{\ourbm}{ManyRef100}

\title{\ourwork: Decoupling Conditioning Cost from References via Compact Residual Conditioning and Spatial Routing}

\author{Wanning He\textsuperscript{*} \quad Yuyao Zhang\textsuperscript{*} \quad Yu-Wing Tai \\
Dartmouth College \qquad {\normalfont\small \textsuperscript{*}Equal contribution; co-first authors.}}

\iclrfinalcopy
\hypersetup{pdftitle={RefRoute: Decoupling Conditioning Cost from References via Compact Residual Conditioning and Spatial Routing},pdfauthor={Wanning He, Yuyao Zhang, Yu-Wing Tai}}
\begin{document}

\maketitle

\CheckSamplingProtocol

\begin{figure}[!ht]
\vspace{-0.2in}
\centering
\includegraphics[width=\linewidth]{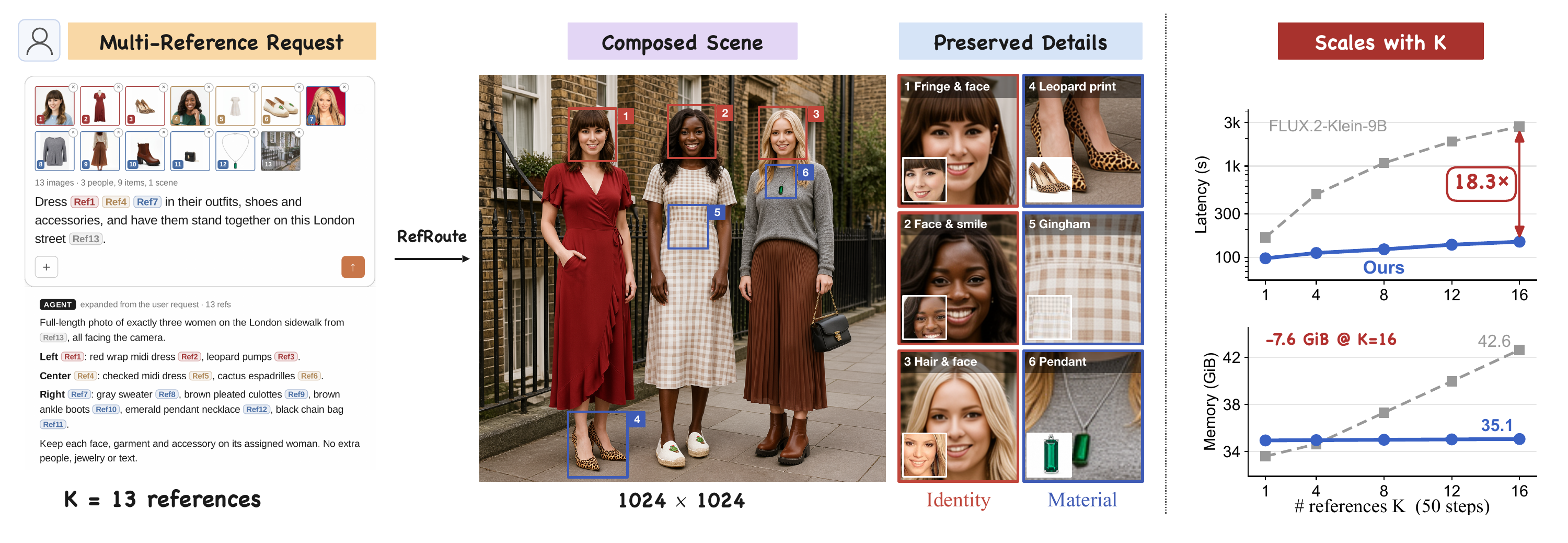}\\
\vspace{-0.2in}
\caption{\textbf{Demo of RefRoute.}
Our method enables efficient many-reference generation while preserving fine appearance details and spatial correspondence.
With 16 references, we achieve an \textbf{18.3$\times$ speedup} over FLUX.2-Klein-9B at 50 steps.}
\label{fig:teaser}
\end{figure}

\begin{abstract}
Multi-reference image generation requires preserving the appearance of multiple subjects while composing them into a coherent scene. However, existing diffusion transformers commonly encode references as dense visual token grids and jointly process them with global attention, making conditioning increasingly expensive as the number and resolution of references grow. We present \textbf{RefRoute}, a framework that addresses both reference representation cost and attention overhead through two complementary mechanisms. \textbf{Compact residual conditioning} combines low-resolution latent tokens with lightweight residual features extracted from full-resolution pixels, reducing reference token counts while retaining fine-grained appearance cues. \textbf{Condition routing and attention routing} align reference tokens with their assigned target regions and restrict cross-reference interactions, while allowing selective reference access beyond region boundaries for scene integration. We further introduce \textbf{RefRoute-Data} for training many-reference generation models and \textbf{ManyRef100}, a benchmark spanning human, object, and mixed compositions with 10--17 references. After many-reference fine-tuning, RefRoute achieves an overall Weighted-Ref-VIEScore of 36.06 on ManyRef100, compared with 8.88 for FLUX.2-Klein-9B. Separate inference-cost evaluations show substantially slower latency growth as the reference count increases: at 16 references, our 50-step and 4-step configurations achieve $18.3\times$ and $14.2\times$ speedups over their corresponding FLUX baselines, respectively. These results establish compact reference representations and spatially routed attention as an effective approach to scalable many-reference image generation.
\end{abstract}

\section{Introduction}
\label{sec:intro}

Recent image generation models increasingly support \emph{multi-reference conditioning}~\citep{esser2024sd3,bfl2025flux2,wu2025qwenimage,batifol2025kontext,xiao2024omnigen,wu2025omnigen2,deng2025bagel,google2025nanobanana}, enabling users to control an image using multiple reference images, layouts, object regions, or other spatial cues~\citep{xiao2023fastcomposer,wang2024msdiffusion,mou2025dreamo,wu2025uno,chen2025xverse,she2025mosaic}. As illustrated in Fig.~\ref{fig:teaser}, a single generation may require several references to preserve the appearance of different entities while placing them at specified locations. This setting raises a fundamental efficiency question: \emph{How should the cost of conditioning scale as the number of references increases?}

Current multimodal diffusion transformers largely answer this question by concatenation. Each reference image is encoded into a dense visual token grid, and the resulting reference tokens are appended to text and noisy target tokens in a unified attention sequence~\citep{tan2024ominicontrol,huang2024iclora,batifol2025kontext,wu2025omnigen2,wei2025mico,peng2026structgen}. If each reference contributes $n$ tokens and the target contains $N_g$ tokens, joint attention has complexity
$O\left((N_g+Kn)^2\right)$,
where $K$ is the number of references. For example, a $1024\times1024$ reference produces 4096 visual tokens, making multiple references ($K > 1$) quickly dominate the input sequence. This reveals \textbf{two distinct sources of unnecessary computation}.

\noindent\textbf{Representation scaling.}
A dense reference representation allocates tokens according to the number and resolution of input references. Yet the information required for generation does not necessarily grow with its references. Simply reducing the reference resolution decreases token count, but can discard precisely the high-frequency cues needed for faithful reconstruction~\citep{medi2025highfreq}. The challenge is therefore not merely to \emph{compress} a reference, but to construct a \emph{bounded representation that preserves generation-relevant appearance information}.

\noindent\textbf{Interaction scaling.}
Even after reducing the representation size, standard concatenation leaves the attention graph dense: every target token can interact with every reference token, and reference tokens from different entities can interact with one another. The model must therefore learn implicitly which reference controls which target region. As $K$ grows, this unrestricted interaction creates unnecessary computation and increases the opportunity for attribute--subject misassociation, identity leakage, and spatial confusion~\citep{chen2026macro,oshima2025multibanana,tian2026multibind,chen2026collapse}. Spatial correspondence should instead be represented explicitly in the computation graph: a reference should primarily communicate with the region it is responsible for, rather than with the entire canvas.

These observations suggest a simple design principle:
\emph{Conditioning should scale with the information and spatial support required by the output, rather than with the resolution and global connectivity of the inputs.}
Achieving this principle requires solving representation and interaction scaling together. A compact representation alone does not prevent all references from attending globally; spatial masking alone does not prevent the reference token count from growing with input resolution. RefRoute addresses both bottlenecks with two coupled mechanisms: \emph{compact residual conditioning} and \emph{condition/attention routing}.

\noindent\textbf{Compact residual conditioning.}
We assign every reference a fixed token budget independent of its resolution. Specifically, a $1024\times1024$ reference is reduced from a $64\times64$ latent grid (4096 tokens) to a $16\times16$ grid (256 tokens). This makes the representation cost independent of the reference resolution. However, the compact latent alone loses fine appearance information. We therefore introduce a lightweight Pixel Residual Compressor that extracts complementary information directly from the original pixels and injects it into the first DiT layer. The resulting representation,
$
H^{\mathrm{ref}} = H^{\mathrm{lr}} + P(r),
$
combines the efficiency of a fixed-size latent representation with a residual pathway for high-frequency appearance cues. %
Because attention complexity depends on the total sequence length, this changes the reference-dependent attention cost by roughly $70\times$ at $K=16$, ignoring text tokens.

\noindent\textbf{Condition and attention routing.}
Compact references still need to be associated with the correct target instances. RefRoute therefore makes this correspondence explicit. We align each reference with its assigned target region by remapping its rotary positional coordinates. Structured attention then blocks direct interactions between distinct references and lets target queries within each assigned region access the corresponding reference. Outside assigned regions, each target query can access the highest-ranked reference, retaining all tokens of that reference. This selective access allows reference information to influence the surrounding scene, supporting effects such as shadows and reflections while limiting global reference interactions.

We introduce \textbf{RefRoute-Data} and \textbf{ManyRef100} to support training and evaluation of many-reference generation. After many-reference fine-tuning, RefRoute achieves an overall Weighted-Ref-VIEScore of 36.06 on ManyRef100, compared with 8.88 for FLUX.2-Klein-9B. Separate inference-cost evaluations show $18.3\times$ and $14.2\times$ speedups at 16 references over the corresponding FLUX baselines at 50 and four sampling steps, respectively. These results demonstrate improved many-reference composition and computational scaling, with a quality trade-off on the predominantly few-reference MICo-Bench.

\noindent\textbf{Contributions.}
Our contributions are threefold:
\begin{itemize}[leftmargin=1.5em, itemsep=0pt, topsep=2pt]
\item We introduce \textbf{compact residual conditioning}, which combines low-resolution latent tokens with pixel-derived residual features to reduce reference token counts while retaining fine-grained appearance cues.

\item We propose \textbf{condition and attention routing}, which makes reference--region correspondence explicit through positional alignment and structured attention, with selective reference access beyond assigned regions.

\item We construct \textbf{RefRoute-Data} with aligned reference--instance--spatial annotations and \textbf{ManyRef100}, a benchmark covering human, object, and mixed compositions with 10--17 references per case, to support training and systematic evaluation of many-reference generation.
\end{itemize}

\section{Related Work}
\label{sec:related_work}

\noindent\textbf{Multi-Reference Controllable Generation.}
Recent diffusion transformers~\citep{peebles2023dit} incorporate visual
conditions by encoding reference images into latent tokens and combining them
with text and noisy target tokens~\citep{tan2024ominicontrol,huang2024iclora,
batifol2025kontext,wu2025uno,mao2025acepp}. Multi-reference methods extend
this design by concatenating multiple reference-token blocks
~\citep{wei2025mico,peng2026structgen,xiao2024omnigen,wu2025omnigen2,
mou2025dreamo,xu2026unicustom}. Other works study combinations of reference
images with spatial layouts, masks, and poses~\citep{xiao2023fastcomposer,
wang2024msdiffusion,mou2025dreamo,wu2025uno,chen2025xverse,she2025mosaic}.

Most of these methods use a shared attention context for the reference and
target tokens. Thus, the token count grows with the number and resolution of
references, while the model must learn which reference should control each
target instance. This can cause attribute--subject misassociation and
appearance confusion as the number of references increases
~\citep{chen2026macro,oshima2025multibanana,tian2026multibind,chen2026collapse}.
StructGen reduces this ambiguity by assigning explicit identifiers to
references and their relations~\citep{peng2026structgen}, but the reference
tokens still form a shared attention context. RefRoute instead explicitly
binds each reference to its designated target region and restricts
unnecessary interactions via condition and attention routing.

\noindent\textbf{Efficient Conditioning and Diffusion Acceleration.}
Several works reduce the cost of processing visual conditions. OminiControl2~\citep{tan2025ominicontrol2}
uses compact condition representations and reuses condition features with asymmetric attention, while
other methods also study efficient conditional processing
~\citep{liu2025nanocontrol}. A separate line of work reduces the number of
denoising evaluations through progressive distillation, consistency training,
and distribution matching~\citep{salimans2022progressive,song2023consistency,
yin2023dmd,sun2026tbsm}. These approaches address condition processing or
sampling cost, but do not specifically address how the conditioning cost
scales with a large number of appearance references. RefRoute focuses on this
many-reference scaling problem through fixed-size per-reference
representations and spatially restricted interactions.

\noindent\textbf{Reference Datasets and Benchmarks.}
Existing benchmarks have expanded from single-subject personalization to
multi-reference and long-context generation. DreamBench~\citep{ruiz2022dreambooth} evaluates subject
preservation and prompt alignment, while
DreamBench++~\citep{peng2024dreambenchpp} improves evaluation with human judgments. OmniContext~\citep{wu2025omnigen2} evaluates in-context generation
with multi-references, MultiRef~\citep{chen2025multiref} combines real
and synthetic multi-condition cases, and MICo-Bench~\citep{wei2025mico}
evaluates multi-image composition with a reference-aware VLM metric. More recent benchmarks increase the number and
diversity of references, including MacroBench~\citep{chen2026macro},
MultiBanana~\citep{oshima2025multibanana}, and StructGen Bench
~\citep{peng2026structgen}. Other works further study multi-human identity
preservation and reference association~\citep{borse2025multihuman,
tian2026multibind,chen2026collapse}.

These benchmarks evaluate reference fidelity, composition, and reference
association, but
do not combine large reference counts with
instance-level spatial assignment and computational scaling. RefRoute-Data
and ManyRef100 target this setting by providing aligned
reference--instance--spatial conditions and evaluating both generation
quality and scaling with the number of references.
\begin{figure}[t]
    \centering
    \includegraphics[width=1.0\linewidth]{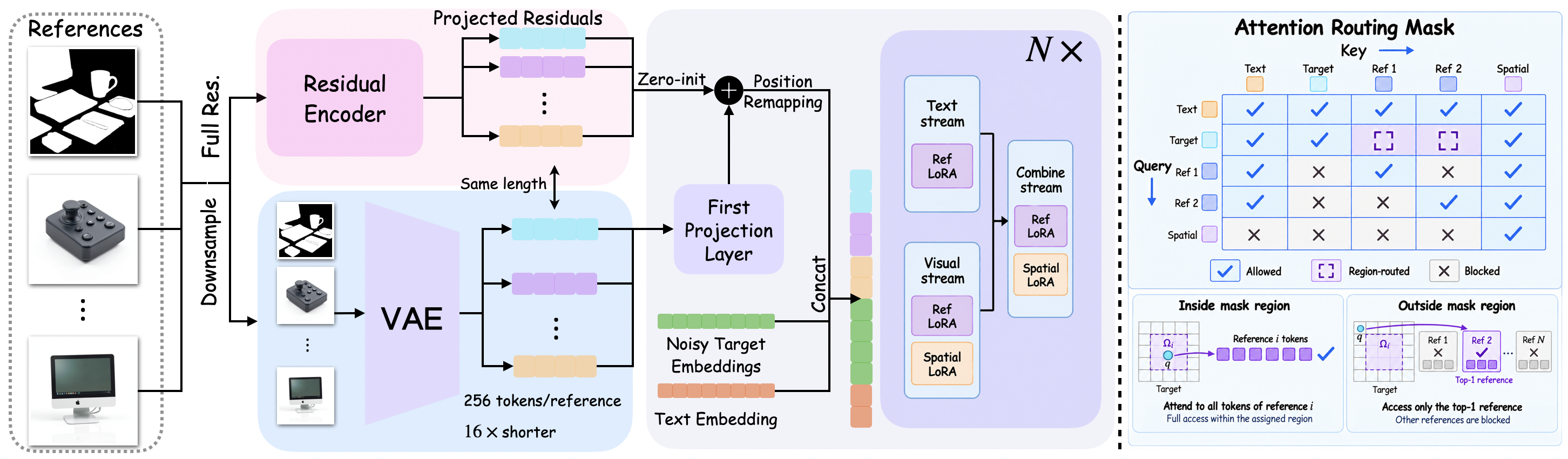}\\
    \vspace{-0.15in}
    \caption{Overview of \ourwork. Each reference is downsampled and VAE-encoded into compact tokens, while a residual encoder extracts complementary features from the full-resolution image. These features are added to the corresponding reference embeddings after the DiT input projection. The augmented embeddings are jointly processed with text and noisy target embeddings through condition-specific adapters and attention routing.}
    \vspace{-0.15in}
    \label{fig:pipeline}
\end{figure}

\section{Method}
We present \ourwork, an efficient framework for multi-reference conditioned generation. Figure~\ref{fig:pipeline} illustrates our compact residual reference representation and routing mechanism within a shared DiT. In the following sections, we first introduce the multi-reference conditioning formulation (Section~\ref{preliminary}), then describe compact residual reference encoding (Section~\ref{effi-cond}) and condition and attention routing (Section~\ref{unif-cond}), and finally present the construction of our training dataset (Section~\ref{training-data}).

\subsection{Preliminaries of Multi-Reference Conditioned Generation}
\label{preliminary}

Given a text prompt $\mathbf{c}$, a target image $\mathbf{x}_0$, and reference images $\{\mathbf{r}_i\}_{i=1}^{N}$, we encode the images into latent representations using a VAE~\citep{rombach2022ldm}. Let $\mathbf{z}_0$ denote the clean target latent and $\mathbf{z}_t=(1-t)\mathbf{z}_0+t\boldsymbol{\epsilon}$ its noisy state at timestep $t$, where $\boldsymbol{\epsilon}\sim\mathcal{N}(\mathbf{0},\mathbf{I})$. We denote the text embeddings by $\mathbf{T}$, the noisy target embeddings by $\mathbf{X}$, and the embeddings of reference $\mathbf{r}_i$ by $\mathbf{C}_r^i$, with the dependence of $\mathbf{X}$ on $t$ omitted for brevity. These embeddings are concatenated into a unified sequence
$[\mathbf{T};\mathbf{X};\mathbf{C}_r^1;\ldots;\mathbf{C}_r^N]$,
and jointly processed by a diffusion transformer (DiT).

To distinguish tokens from different images, each visual token is assigned a coordinate $(\tau_i,h,w)$, where $(h,w)$ denotes its spatial position and $\tau_i$ identifies its source image. These coordinates are incorporated through rotary positional embeddings (RoPE)~\citep{su2024roformer}.

The model is optimized with the standard flow-matching objective~\citep{liu2023rectifiedflow}:
\begin{equation}
\mathcal{L}_{\mathrm{FM}}=\mathbb{E}\left[\left\|\mathbf{v}_{\theta}(\mathbf{z}_t,t,\mathbf{c},\{\mathbf{r}_i\}_{i=1}^{N})-(\boldsymbol{\epsilon}-\mathbf{z}_0)\right\|_2^2\right],
\end{equation}
where $\mathbf{v}_{\theta}$ denotes the predicted velocity. Although this unified formulation is flexible, its sequence length grows with the reference token counts, leading to substantial attention overhead. Moreover, unrestricted joint attention allows information from different references to mix across target regions. These challenges motivate our compact reference representation and explicit attention routing.

\subsection{Compact Residual Reference Representation}
\label{effi-cond}

The first bottleneck is the number of reference tokens. A $\TargetResolution\times\TargetResolution$ reference produces a $\NativeGridSide\times\NativeGridSide$ latent grid with $\NativeRefTokens$ tokens, yielding $\IllustrativeNativeTokens$ reference tokens for $\IllustrativeReferences$ references. We reduce this cost by downsampling each reference by a factor of $\ReferenceDownsample$ before VAE encoding. This produces a $\CompactGridSide\times\CompactGridSide$ grid with $\CompactRefTokens$ tokens per reference, which we use in our experiments with up to 16 references.

Downsampling, however, removes fine-grained appearance information. We therefore introduce a lightweight \emph{Pixel Residual Compressor} that extracts complementary features from the original full-resolution reference. As shown in Figure~\ref{fig:pipeline}, each reference follows two parallel pathways: a compact VAE pathway that provides low-resolution latent tokens, and a pixel pathway that produces a residual feature sequence with the same spatial grid and token count.

Let $\mathbf{Z}_{\mathrm{lr}}^i=\mathcal{E}(\operatorname{Downsample}(\mathbf{r}_i))$ denote the compact latent tokens of reference $\mathbf{r}_i$, where $\mathcal{E}$ is the VAE encoder. The DiT input projection maps these tokens to the transformer hidden dimension. In parallel, the Pixel Residual Compressor $P$ extracts features from $\mathbf{r}_i$, which are mapped to the same hidden dimension by a zero-initialized projection. The resulting reference embeddings are
\begin{equation}
\mathbf{C}_r^i=\operatorname{Proj}_{\mathrm{in}}(\mathbf{Z}_{\mathrm{lr}}^i)+\operatorname{Proj}_{0}(P(\mathbf{r}_i)),
\end{equation}
where $\operatorname{Proj}_{\mathrm{in}}$ is the pretrained DiT input projection and $\operatorname{Proj}_{0}$ is a trainable projection whose weights and bias are initialized to zero. The residual is added elementwise to the corresponding projected reference tokens, once before the first DiT block. This preserves the compact sequence length while allowing full-resolution appearance information to complement the low-resolution representation. %
Architectural details are provided in Appendix~\ref{sec:appendix:compressor}.

\noindent\textbf{Pretraining the Pixel Residual Compressor.}
We pretrain the compressor to preserve appearance information useful for generation. Rather than encouraging a unified representation to retain all reference information~\citep{pfp2025}, we process each reference independently and combine reconstruction with subject-driven generation.

At each training iteration, we sample either a reconstruction task, whose target is the reference itself, or a subject-driven generation task, whose target depicts the same subject under a different pose, viewpoint, or context. For a reference image $\mathbf{r}$, the target--reference pair is
\begin{equation}
(\mathbf{x}_0,\mathbf{r})=(\mathbf{r},\mathbf{r})\quad\text{or}\quad(\mathbf{x}_{\mathrm{subj}(\mathbf{r})},\mathbf{r}),
\end{equation}
where $\mathbf{x}_{\mathrm{subj}(\mathbf{r})}$ depicts the same subject as $\mathbf{r}$. Both tasks use the flow-matching objective from Section~\ref{preliminary}. Reconstruction encourages faithful appearance encoding, while subject-driven generation encourages transferable identity and appearance cues. For spatially conditioned examples, the compressor and generation model are trained to reconstruct the corresponding target image. The pretraining lays the foundation for unified multi-reference training.

\subsection{Condition and Attention Routing}
\label{unif-cond}

Compact encoding reduces the reference token count, but does not control how information from different references is combined during generation. We therefore introduce condition-specific adaptation and positional remapping, together with an attention routing mask that regulates interactions between references and target regions.

\noindent\textbf{Condition Routing.}
Each reference image $\mathbf{r}_i$ is encoded into compact reference embeddings $\mathbf{C}_r^i$ using the two pathways described in Section~\ref{effi-cond}. These embeddings are processed together with the text embeddings $\mathbf{T}$, noisy target embeddings $\mathbf{X}$, and spatial condition embeddings $\mathbf{C}_s$ when available. We use separate reference and spatial LoRA adapters~\citep{hu2022lora} for condition-specific adaptation. As illustrated in Figure~\ref{fig:pipeline}, the reference adapter operates in both the text and visual streams, while the spatial adapter operates in the visual stream; both are used in the subsequent combined stream.

To account for the reduced spatial resolution of compact references, we remap their rotary positional coordinates. For a reference token at grid position $(h,w)$, we use
$(\tau_i,h,w)\longmapsto(\tau_i,sh,sw)$, where $s$ is the reference downsampling factor and $\tau_i$ identifies the reference image. This rescaling expresses reference positions at the spatial scale of the target latent grid. The association between each reference and its designated target region is enforced by the attention routing mask below.

\noindent\textbf{Attention Routing.}
Unrestricted joint attention allows references to exchange information with one another and influence unrelated target regions. We regulate these interactions using the routing structure shown in Figure~\ref{fig:pipeline}. Following condition-specific adaptation, the queries, keys, and values are organized as:
$\mathbf{U}=[\mathbf{U}_{\mathrm{T}};\mathbf{U}_{X};\mathbf{U}_{C_r^1};\ldots;\mathbf{U}_{C_r^N};\mathbf{U}_{C_s}],\ \mathbf{U}\in\{\mathbf{Q},\mathbf{K},\mathbf{V}\},$
where the spatial condition component is omitted when unavailable. An additive routing bias $\mathbf{M}$ controls attention:
\begin{equation}
\operatorname{Attn}(\mathbf{Q},\mathbf{K},\mathbf{V})=\operatorname{Softmax}\left(\frac{\mathbf{Q}\mathbf{K}^{\top}}{\sqrt{d_k}}+\mathbf{M}\right)\mathbf{V},
\end{equation}
where $d_k$ is the key dimension. Each entry of $\mathbf{M}$ is zero for a permitted query--key connection and $-\infty$ for a prohibited connection.

The mask first separates the reference streams: reference queries can attend to their own reference tokens, text tokens, and spatial condition tokens, but cannot attend to noisy target tokens or other references. Text queries retain access to all token groups, while target queries retain access to text, target, and spatial condition tokens. Spatial condition queries attend only to spatial condition tokens. These rules preserve shared textual and spatial guidance while preventing direct information exchange between different references.

Target-to-reference attention is further controlled by spatial assignments. Let $\Omega_i$ denote the target-token region assigned to reference $\mathbf{r}_i$. A target query within $\Omega_i$ can attend to all tokens in $\mathbf{C}_r^i$, providing full access to the corresponding reference appearance within its designated region.

Strictly blocking reference information outside the assigned regions can limit effects that extend beyond object boundaries, such as shadows, illumination, and contact. We therefore allow target queries outside the assigned regions to access only the top-ranked reference. All tokens of the selected reference remain accessible, while connections to the other references are blocked. This selection operates at the reference level rather than the individual-token level. Together, these rules maintain explicit reference--region correspondence while permitting limited reference influence outside the assigned regions.

\noindent\textbf{Overall Training.}
We train \ourwork{} in two stages. In Stage 1, we pretrain the Pixel Residual Compressor using the reconstruction and subject-driven generation tasks described in Section~\ref{effi-cond}. In Stage 2, we initialize the compressor from the pretrained checkpoint and train the model with multiple references and spatial conditions using the condition and attention routing mechanisms above. Both stages use the flow-matching objective defined in Section~\ref{preliminary}.

\subsection{Dataset Creation}
\label{training-data}

Existing multi-reference datasets~\citep{ye2025echo4o,liu2026opensubject}
provide broad task coverage, but MICo~\citep{wei2025mico}, for example, has
three limitations: 89.2\% of MICo-Bench cases have at most five references,
references lack instance-level spatial annotations, and human references
often mismatch the target identity. \ourdata{} provides per-instance
references, boxes, and masks for 20k object scenes (143k instances) and
1k human scenes (Appendix~\ref{sec:appendix:data-construction}); its mean
reference--target ArcFace similarity is 0.53, versus 0.19 for the
MICo-150K human subset.

\begin{table}[t]
\centering
\vspace{-0.05in}
\color{black}
\caption{{\textbf{MICo-Bench results.} Completed evaluations cover all 897 cases at native $1024\times1024$ resolution. Columns report category-level and overall Weighted-Ref-VIEScore ($0$--$100$, higher is better). Rows above the first rule are reported results; FLUX.2 rows and ours are evaluated by us.}}
\label{tab:mico}
\footnotesize
\setlength{\tabcolsep}{3pt}
\renewcommand{\arraystretch}{1.08}
\begin{tabular*}{\linewidth}{@{\extracolsep{\fill}}lrrrrr@{}}
\toprule
Method & Object & Person & HOI & De\&Re & Overall \\
\midrule
Gemini-3-Pro-Image-Preview & 50.59 & \underline{54.75} & 50.21 & 52.13 & 51.76 \\
GPT-Image-1.5 & \textbf{56.66} & 46.16 & 52.35 & 48.46 & 50.60 \\
Gemini-2.5-Flash-Image & 48.01 & 41.79 & 49.64 & 49.44 & 47.83 \\
Qwen-Image-MICo & 52.38 & 21.11 & 34.95 & 37.42 & 35.86 \\
BAGEL-MICo & 38.98 & 28.45 & 25.30 & 44.51 & 34.41 \\
OmniGen2-MICo & 46.26 & 22.85 & 32.18 & 36.82 & 33.82 \\
Qwen-Image-Edit-2509 & 39.77 & 20.23 & 19.95 & 29.92 & 27.47 \\
BLIP3o-Next-MICo & 40.31 & 11.41 & 24.97 & 26.23 & 25.21 \\
Lumina-Dimoo-MICo & 38.44 & 12.14 & 24.66 & 21.32 & 23.32 \\
FLUX.1-Kontext-dev ($512^2$) & 21.40 & 14.33 & 12.67 & 7.24 & 12.51 \\
\midrule
{FLUX.2-Klein-Base-9B} & \underline{{54.66}} & \textbf{{55.14}} & \underline{{52.44}} & \underline{{59.56}} & \textbf{{55.67}} \\
{FLUX.2-Klein-9B} & {50.63} & {53.24} & {51.07} & \textbf{{62.36}} & \underline{{55.18}} \\

\midrule
{Ours-MICo} & {48.22} & {49.65} & \textbf{{52.81}} & {56.68} & {52.81} \\
{Ours-MICo-Distilled} & {46.72} & {43.93} & {49.88} & {54.80} & {49.93} \\
\bottomrule
\end{tabular*}
\vspace{-0.2in}
\end{table}

\section{Experiments}
\label{sec:experiments}

\noindent\textbf{Experimental Setup.}
\label{sec:exp:impl}
\ourwork{} is built on FLUX.2-Klein-9B, and our evaluations use
$1024\times1024$ outputs in BF16.
We evaluate on two complementary benchmarks:
MICo-Bench~\citep{wei2025mico}, where 89.2\% of 897 cases contain
two to five references, and our \ourbm{}, comprising 100 cases
(30 Human, 30 Object, and 40 Mixed) with 10--17 references
and instance-level spatial annotations.
We measure generation quality using Weighted-Ref-VIEScore,
which combines reference fidelity, semantic consistency,
and perceptual quality.
Data construction, model configurations, and evaluation protocols
are detailed in Appendices~\ref{sec:appendix:data-construction}
and~\ref{sec:appendix:max-reproducibility}.

\noindent\textbf{Multi-Reference Composition.}
\label{sec:exp:quant}
We first evaluate our MICo variants before additional
many-reference fine-tuning.
As shown in Table~\ref{tab:mico}, the undistilled and distilled
variants achieve overall scores of 52.81 and 49.93, respectively.
The undistilled variant surpasses the reported scores of
Gemini-3-Pro-Image-Preview (51.76) and GPT-Image-1.5 (50.60).
Both variants score below their corresponding FLUX baselines
(55.67 and 55.18), while offering substantial computational
savings as the reference count increases, as shown next.

\noindent\textit{Scaling with the Number of References.}
\label{sec:exp:efficiency}
We examine both inference cost and generation quality
as the number of references increases.

\begin{figure}[t]
\centering
\includegraphics[width=\linewidth]{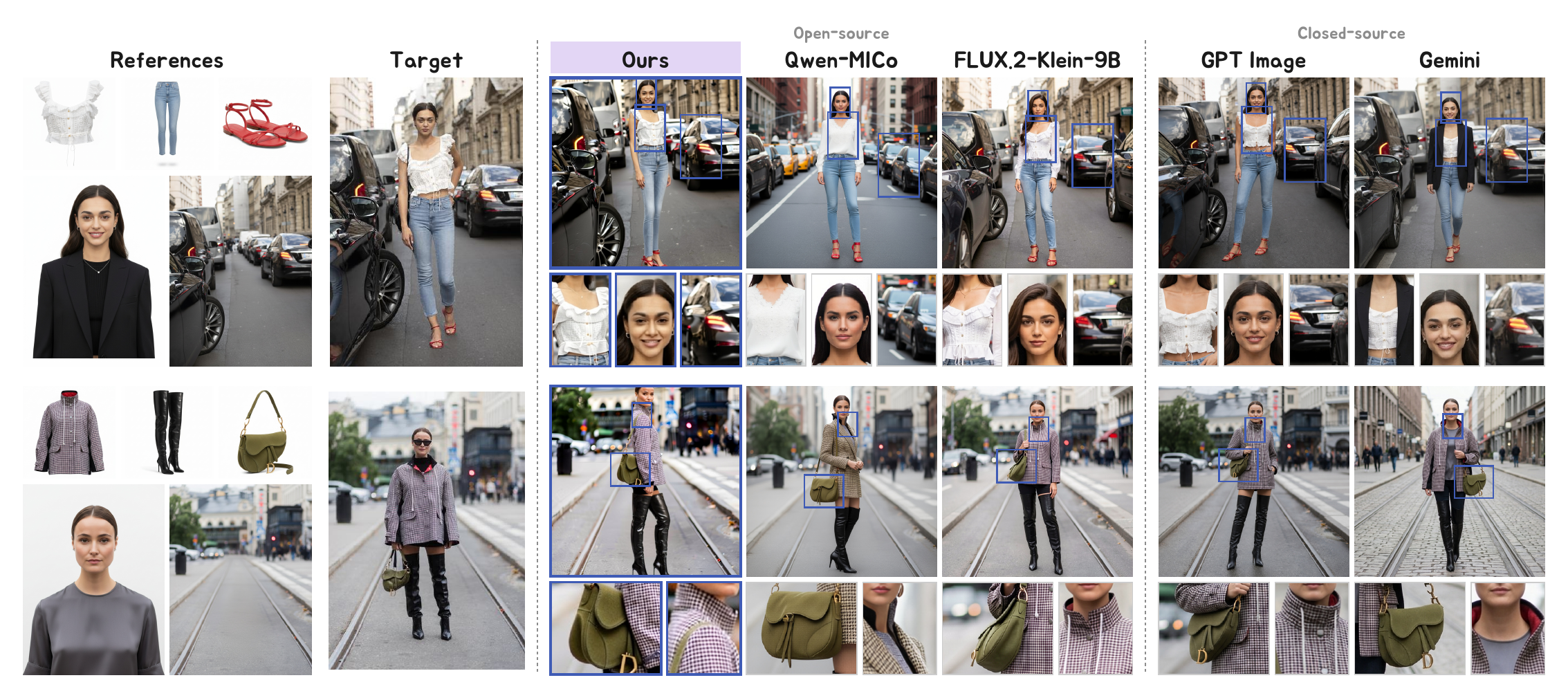}\\
\vspace{-0.1in}
\includegraphics[width=\linewidth]{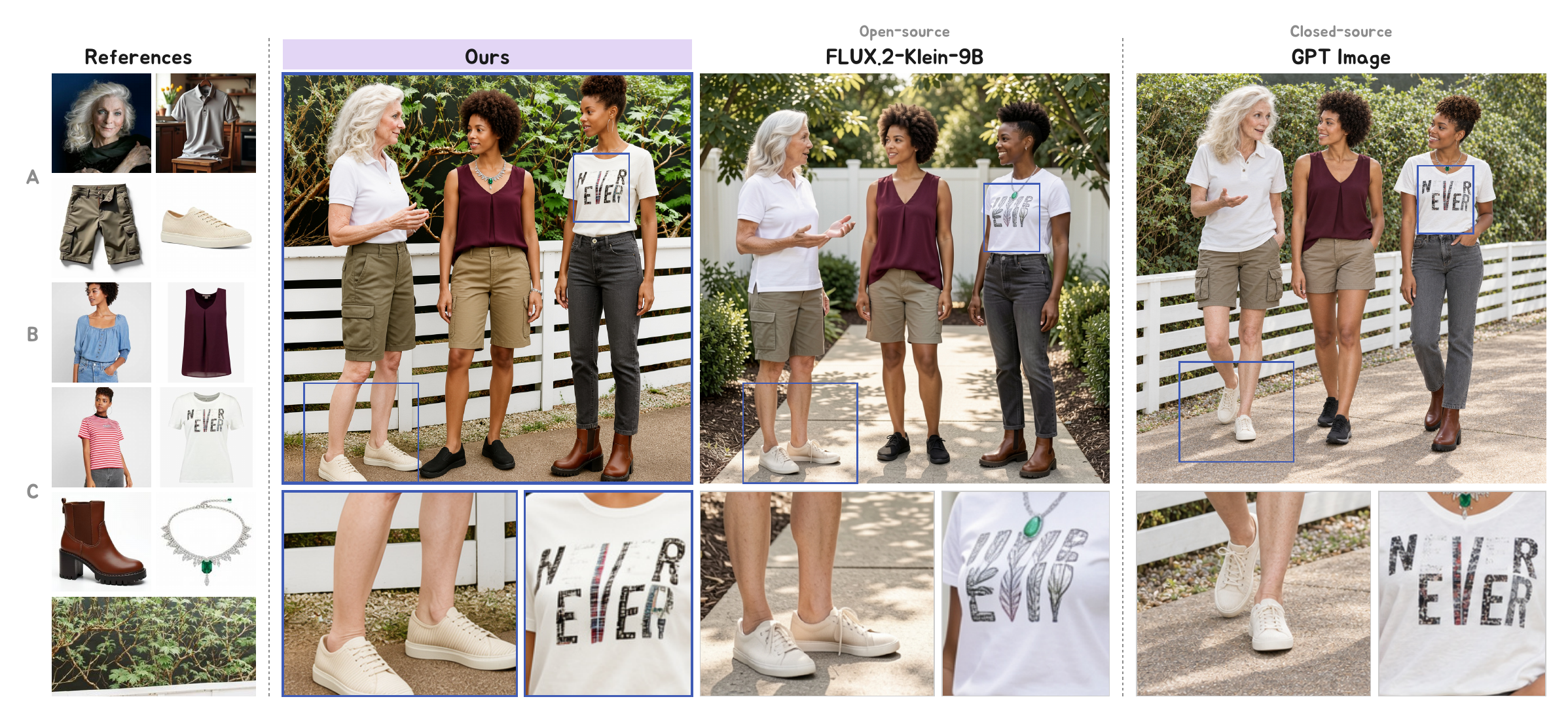}\\
\vspace{-0.2in}
\caption{\textbf{Qualitative comparison.} 
Top: MICo-Bench, with few references; columns show the references, the target, \ourwork{}, open-source baselines (Qwen-Image-MICo, FLUX.2-Klein-9B), and closed-source systems (GPT-Image, Gemini). Bottom: \ourbm{}, with 10--17
references and assigned regions overlaid.}
\vspace{-0.2in}
\label{fig:qual}
\end{figure}
\begin{table}[t]
\centering
\color{black}
\caption{Comparison on \ourbm{}. We evaluate $1024\times1024$ generation with many references on our benchmark. Ours-ManyRef achieves the best overall and per-category scores; without many-reference fine-tuning, Ours-MICo outperforms both FLUX.2 baselines.}
\label{tab:rrbench}
\small
\setlength{\tabcolsep}{2pt}
\renewcommand{\arraystretch}{1.08}
\begin{tabular*}{\linewidth}{@{\extracolsep{\fill}}lrrrrrrr@{}}
\toprule
Method & Overall & Human & Object & Mixed & $W$ & $SC$ & $PQ$ \\
\midrule
{FLUX.2-Klein-Base-9B} & 4.18 & 10.79 & 0.09 & 2.28 & 0.268 & 1.470 & 6.949 \\
{FLUX.2-Klein-9B} & 8.88 & \underline{17.07} & 4.56 & 5.98 & 0.375 & 2.679 & 7.319 \\
{GPT-Image-1} & \underline{27.92} & {10.31} & \underline{55.95} & \underline{20.09} & 0.574 & \underline{5.613} & \textbf{7.763} \\
\midrule
{Ours-MICo} & 17.32 & 13.60 & 36.10 & 6.02 & \underline{0.639} & 3.758 & 6.171 \\
{Ours-ManyRef} & \textbf{36.06} & \textbf{28.10} & \textbf{57.37} & \textbf{26.04} & \textbf{0.709} & \textbf{6.428} & \underline{7.583} \\

\bottomrule
\end{tabular*}
\vspace{3pt}
\begin{minipage}{\linewidth}
\scriptsize\color{black}
\end{minipage}
\vspace{-0.2in}
\end{table}

\noindent\textit{Qualitative Comparison.}
\label{sec:exp:qual} Figure~\ref{fig:teaser} demonstrates many-reference generation with RefRoute, combining 13 references into a coherent scene while preserving facial identity and clothing patterns. The highlighted details include leopard-print shoes, a gingham dress, and a pendant, illustrating the retention of fine-grained appearance cues alongside correct reference--subject associations. Figure~\ref{fig:qual} presents qualitative comparisons on MICo-Bench. In the first row, RefRoute more faithfully preserves the woman's facial identity, the structure of her top, and the background. In comparison, Qwen-MICo and Gemini alter the referenced top, while Qwen-MICo and FLUX.2-Klein-9B show discrepancies in facial identity. In the second row, RefRoute better retains the bag's shape and construction details, as well as the jacket's checkered pattern. The \ourbm{} example demonstrates our better results on the t-shirt logo and shoe details. More comparisons are in the appendix.

\FloatBarrier

\begin{figure}[t]
\centering
\includegraphics[width=\linewidth]{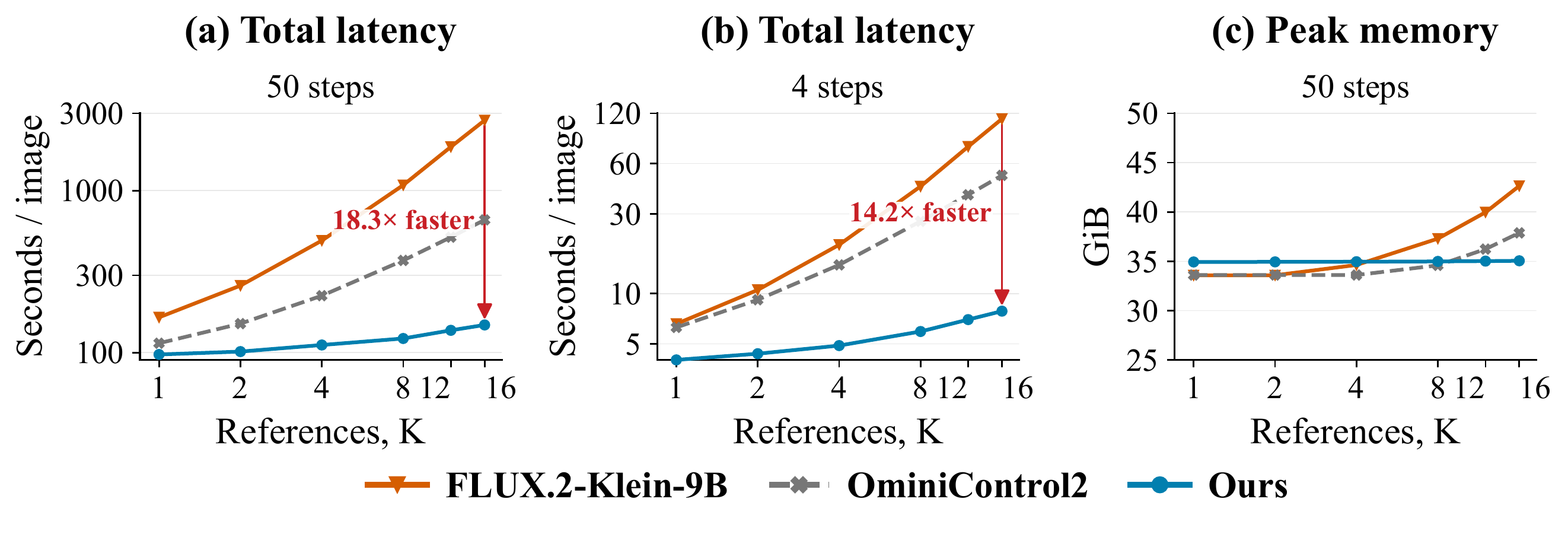}\\
\vspace{-0.22in}
\caption{\textbf{Inference cost versus reference count.}
Compared with FLUX.2-Klein baselines and an OminiControl2
reference-branch implementation, \ourwork{} exhibits slower
latency growth and nearly constant peak allocated memory
across the measured configurations.}
\vspace{-0.15in}
\label{fig:scaling}
\end{figure}
\begin{table}[t]
\centering
\caption{\textbf{Residual source and injection space.}
Left: Stage~1 reconstruction on 100 images.
Right: \ourbm{} after a matched Stage~2 budget of 10k steps. Human and mixed cells are ArcFace face cosine / mIoU / RRBA;
object cells are DINO / mIoU / RRBA.
Bold is best in each panel.}
\label{tab:residual}
\scriptsize
\setlength{\tabcolsep}{2.2pt}
\renewcommand{\arraystretch}{1.15}
\resizebox{\linewidth}{!}{%
\begin{tabular}{@{}c@{\hspace{10pt}}c@{}}
\textbf{(a) Reconstruction} & \textbf{(b) \ourbm{}}\vspace{-0.1in} \\
[0.35em]
\begin{tabular}[t]{@{}lcccc@{}}
\toprule
Design & PSNR$\uparrow$ & SSIM$\uparrow$ & DISTS$\downarrow$ & HF-MAE$\downarrow$ \\
\midrule
VAE latent $\to$ DiT & 24.22 & 0.725 & 0.0837 & 0.0220 \\
Pixel $\to$ VAE & 23.57 & 0.724 & 0.0949 & 0.0226 \\
\textbf{Pixel $\to$ DiT} & \textbf{24.78} & \textbf{0.745} & \textbf{0.0789} & \textbf{0.0207} \\
\bottomrule
\end{tabular}
&
\begin{tabular}[t]{@{}lccc@{}}
\toprule
Design & Human & Mixed & Object \\
\midrule
Concat & 0.264 / 0.177 / 0.390 & 0.217 / 0.177 / 0.385 & 0.701 / 0.426 / 0.794 \\
Pixel $\to$ VAE & 0.254 / 0.168 / 0.413 & 0.223 / 0.198 / 0.427 & 0.696 / 0.409 / 0.799 \\
\textbf{Pixel $\to$ DiT} & \textbf{0.300 / 0.217 / 0.517} & \textbf{0.248 / 0.221 / 0.467} & \textbf{0.731 / 0.444 / 0.824} \\
\bottomrule
\end{tabular}
\end{tabular}%
}
\vspace{-0.2in}
\end{table}

\noindent\textit{Computational cost.}
Figure~\ref{fig:scaling} reports inference cost for $K=1$ to $16$.
At $K=16$, \ourwork{} achieves $18.3\times$ and $14.2\times$
speedups over the corresponding FLUX baselines at 50 and
4 sampling steps, respectively.
Across this range, peak allocated memory at 50 steps increases
by only $0.12\,\mathrm{GiB}$, compared with
$9.05\,\mathrm{GiB}$ for the dense baseline.
We also obtain $4.5\times$ and $6.5\times$ speedups over an
OminiControl2 reference-branch implementation on the same backbone.
This comparison measures mechanism cost rather than the
complete trained OminiControl2 system.
Measurement settings are provided in
Appendix~\ref{sec:appendix:max-reproducibility}.

\noindent\textbf{Many-reference quality.}
Table~\ref{tab:rrbench} reports results on \ourbm{}.
Before additional many-reference fine-tuning, our MICo variant
achieves an overall score of 17.32, exceeding
FLUX.2-Klein-Base-9B (4.18) and FLUX.2-Klein-9B (8.88).
Further fine-tuning raises the score to 36.06 and achieves
the best results across Human, Object, and Mixed categories.

\FloatBarrier
\providecommand{\tbd}[1]{\textcolor{red}{TBD (#1)}}
\section{Ablation Study}
\label{sec:ablation}
We ablate the compact residual representation and dynamic attention
routing. Training configurations are provided in
Appendix~\ref{sec:appendix:training}.
\subsection{Residual Representation Design}
\label{sec:abl:residual}

Compressing references to a
$\CompactGridSide\times\CompactGridSide$ grid can discard
fine appearance details. Table~\ref{tab:residual} examines
how residual source and injection space affect their recovery.
On Stage~1 reconstruction, pixel-to-DiT achieves the best results
across all four metrics, reaching 24.78\,dB PSNR compared with
23.57\,dB for pixel-to-VAE and 24.22\,dB for VAE-latent-to-DiT.

On \ourbm{}, with a matched Stage~2 budget of 10k steps and
routing enabled, pixel-to-DiT performs best across all categories
and metrics. Compared with concatenation without a residual,
it improves Human face similarity from 0.264 to 0.300 and
Object DINO similarity from 0.701 to 0.731.
Human RRBA also increases from 0.390 to 0.517, indicating
better reference--region binding.
These results support extracting residual features from pixels
and injecting them into DiT space.

\begin{wrapfigure}{r}{0.5\textwidth}
    \centering
    \vspace{-0.1in}
    \includegraphics[width=\linewidth]{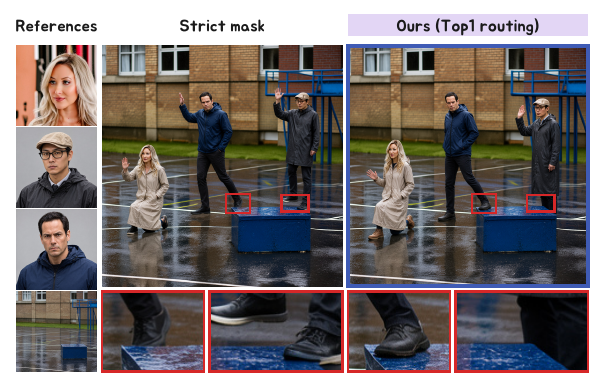}\\
    \vspace{-0.2in}
    \caption{Strict spatial masking vs. dynamic top-1 routing with the same checkpoint, seed, and prompt. Insets highlight reflections and ground contact outside assigned regions.}
    \label{fig:abl_routing}
     \vspace{-0.10in}
\end{wrapfigure}

\subsection{Condition Routing Design}
Restricting reference access to assigned regions can limit
interactions with the surrounding scene, such as shadows
and reflections. Our dynamic top-K (empirically set K=1) mechanism allows each target query outside these regions to attend to the highest-ranked reference, while retaining access to all tokens of that reference.
We compare this mechanism with a strict spatial mask that
blocks reference access outside assigned regions.
The qualitative comparison in Figure~\ref{fig:abl_routing} examines whether dynamic selection improves reference--environment interactions while preserving reference appearance. With the strict mask, the man appears to float above the ground, whereas dynamic routing preserves contact between his feet and the ground. 

\begin{wraptable}{r}{0.5\textwidth}
    \centering
    \vspace{-0.15in}
    \caption{\textbf{Effect of Stage~1 pretraining.}
    Person-count MAE after a matched Stage~2 budget of 10k steps.
    Lower is better; best results are in bold.}
    \label{tab:stage1}
    \small
    \setlength{\tabcolsep}{4pt}
    \begin{tabular}{@{}lccc@{}}
        \toprule
        & \multicolumn{3}{c}{Count MAE$\downarrow$} \\
        \cmidrule(lr){2-4}
        Setting & Overall & Human & Mix \\
        \midrule
        w/o Stage~1 & 1.643 & 1.300 & 1.900 \\
        \textbf{w/ Stage~1}
        & \textbf{1.057} & \textbf{1.033} & \textbf{1.075} \\
        \bottomrule
    \end{tabular}
    \vspace{-0.15in}
\end{wraptable}
\paragraph{Effect of Stage~1 Pretraining.}
We ablate Stage~1 pretraining by training Stage~2 with and without the pretrained compressor for a matched budget of 10k steps. On scenes requesting ten people, pretraining reduces person-count MAE from $1.643$ to $1.057$, with the largest improvement on mixed scenes ($1.900$ to $1.075$; Table~\ref{tab:stage1}). These results suggest that learning reference representations separately facilitates subsequent multi-reference binding under a fixed training budget. Qualitative examples in the appendix (Figure~\ref{fig:stage1}) further illustrate the preservation of facial identity and fine-grained appearance details after Stage~1.

\noindent\textbf{Limitations.}
Our current study focuses on many-reference image conditioning, where the conditioning inputs are visual references. Although RefRoute substantially reduces the representation and attention costs associated with increasing numbers of reference images, we do not investigate settings that involve long text conditioning, such as detailed textual descriptions or combinations of long text prompts with many visual references. Such settings introduce additional challenges in representing and routing long-range semantic information, which are beyond the scope of the current framework. Extending RefRoute to jointly handle large numbers of visual references and long textual conditions is an interesting direction for future work.

\FloatBarrier
\section{Conclusion}
\label{sec:conclusion}

We presented \textbf{RefRoute}, a framework for scalable many-reference image generation that addresses both reference representation cost and attention overhead. Compact residual conditioning preserves fine-grained appearance cues within a reduced token budget, while condition and attention routing spatially align references with their target regions and suppress unnecessary cross-reference interactions. We further introduced \textbf{RefRoute-Data} and \textbf{ManyRef100} to enable training and systematic evaluation in the many-reference regime. Experiments demonstrate substantially improved composition quality on ManyRef100 and more favorable inference-cost scaling as the number of references increases. At the same time, the lower performance on the predominantly few-reference MICo-Bench highlights the challenge of maintaining strong generation quality across different reference-count regimes. Overall, our results show that compact reference representations and spatially structured attention provide complementary mechanisms for scaling multi-reference generation, and suggest a broader direction toward conditioning architectures whose computational capacity adapts to the complexity of the desired scene.

\newpage
\subsection*{AI use statement}

In this work, we used generative AI tools in three ways. \textbf{(1) Synthetic data:} the target images and part of the references in \ourdata{} and \ourbm{} were synthesized with GPT, HiDream-I1-Full, FLUX.2-klein-4B, and InfiniteYou (Appendix~\ref{sec:appendix:data-construction}). \textbf{(2) Data filtering:} vision-language models (Qwen3-VL and GPT-based judges) were used to verify, relabel, and screen generated samples during data construction (Appendix~\ref{sec:appendix:data-construction}). \textbf{(3) Evaluation:} generation quality is scored with Weighted-Ref-VIEScore, which relies on multimodal LLM judges (GPT-5.4 for semantic consistency and perceptual quality, and Qwen3-VL-30B-A3B-Instruct for the reference weights; Appendix~\ref{sec:appendix:max-reproducibility}). In addition, we used generative AI tools to draft parts of this paper and to edit it for readability. The research questions, method design, implementation, and interpretation of results are the authors' own. Synthesized and filtered images were retained only after the automatic detection, face-matching, and geometric and semantic checks described in Appendix~\ref{sec:appendix:data-construction}, and all AI-assisted text was checked by the authors against the experimental records. We take responsibility for the final content of this work, including text, claims, or artifacts produced with the aid of generative AI.

\subsection*{Ethics statement}
This work does not involve human-subject experiments or the collection of personal information from individuals. The experiments use image references and generated content for research on multi-reference image generation. We take care to use data and pretrained models in accordance with their applicable licenses and usage terms. We acknowledge that image generation systems can be used in applications involving the representation or manipulation of people and other sensitive visual content; such applications are outside the scope of this work and may require additional safeguards.

\subsection*{Reproducibility statement}
We provide detailed descriptions of the RefRoute architecture, compact residual conditioning, condition and attention routing, training procedure, and distillation strategy in the paper and supplementary material. We also report the implementation details, model configurations, computational cost measurements, and evaluation protocols used in our experiments. The construction and evaluation protocols of RefRoute-Data and ManyRef100 are described in detail to facilitate independent evaluation of high-cardinality multi-reference generation. We will release the source code, model configurations, data-generation and evaluation scripts, and the relevant dataset metadata upon publication, subject to the licensing and redistribution requirements of the underlying data and pretrained models.

\newpage \bibliography{iclr2027_conference_final}
\bibliographystyle{iclr2027_conference}

\appendix
\section{Appendix}
\label{sec:appendix}

This appendix follows the experimental claims: data and benchmark construction,
evaluation setup and baseline configurations, implementation and training
details, and inference-cost measurements and auxiliary probes.

\begingroup
\color{black}
\subsection{Data and Benchmark Construction}
\label{sec:appendix:data-construction}

Many-reference generation requires associating each reference with its
corresponding subject in a composed scene.
We construct \ourdata{} for training and \ourbm{} for evaluation.
Each training record pairs a target image and scene description with
per-instance references, bounding boxes, and masks, providing explicit
reference-to-instance correspondence.

\paragraph{Data scale.}
The construction manifests contain 20,000 object scenes with
143,249 retained instances and 1,000 human scenes with 4,123
annotated people and 4,067 exported references.
These totals describe the constructed data; the subset used for
training depends on the checkpoint configuration.
Retained object instances cover the requested inventory rather than
all visible objects, and exported human references may contain
identity uncertainty.

\paragraph{Training data construction.}
We use separate object and human branches within a shared four-stage
pipeline (Figure~\ref{fig:app-pipeline}).

\begin{itemize}[leftmargin=1.5em, itemsep=0pt, topsep=2pt]
    \item \textbf{Prompt generation.}
    For objects, we sample five to ten categories from an LVIS-derived
    vocabulary and compose scene descriptions.
    For humans, we specify scenes with two to eight people, including
    their attributes and poses.

    \item \textbf{Target synthesis.}
    We generate object scenes with HiDream-I1-Full and human scenes
    with GPT at $1024\times1024$ resolution.
    Human scenes use input identity images for two to four people
    and text-only conditioning for five to eight people.

    \item \textbf{Instance annotation.}
    For objects, GroundingDINO localizes the requested categories,
    and one-to-one assignment associates detections with the inventory.
    For humans, person detection, pose estimation, and face matching
    associate individuals with their spatial annotations and available
    identity images.
    SAM produces instance masks in both branches.

    \item \textbf{Verification and reference construction.}
    We filter or relabel unreliable object instances using geometric
    and semantic checks, then generate studio-style references from
    target crops with FLUX.2-klein-4B, retaining the crops when
    generation fails.
    For humans, we reuse an input identity image when its ArcFace
    cosine similarity to the matched target face is at least 0.50;
    otherwise, a reference may be synthesized from the target face.
\end{itemize}
\paragraph{Many-reference benchmark.}
We construct \ourbm{} to evaluate composition with larger reference
sets across human, object, and mixed scenes.
It contains 100 cases: 30 Human, 30 Object, and 40 Mixed, each with
10--17 references and explicit reference-to-instance spatial assignments.
Object references derive from an earlier target version, while human
references are rerendered.
The benchmark therefore supports controlled evaluation of
many-subject composition; generalization from independently
photographed references remains outside its scope.
Representative cases are provided in
Appendix~\ref{sec:appendix:walkthrough}.

\endgroup

\subsection{Worked Examples and Construction Flow}
\label{sec:appendix:walkthrough}

\begin{figure}[htbp]
\centering
\includegraphics[width=\linewidth]{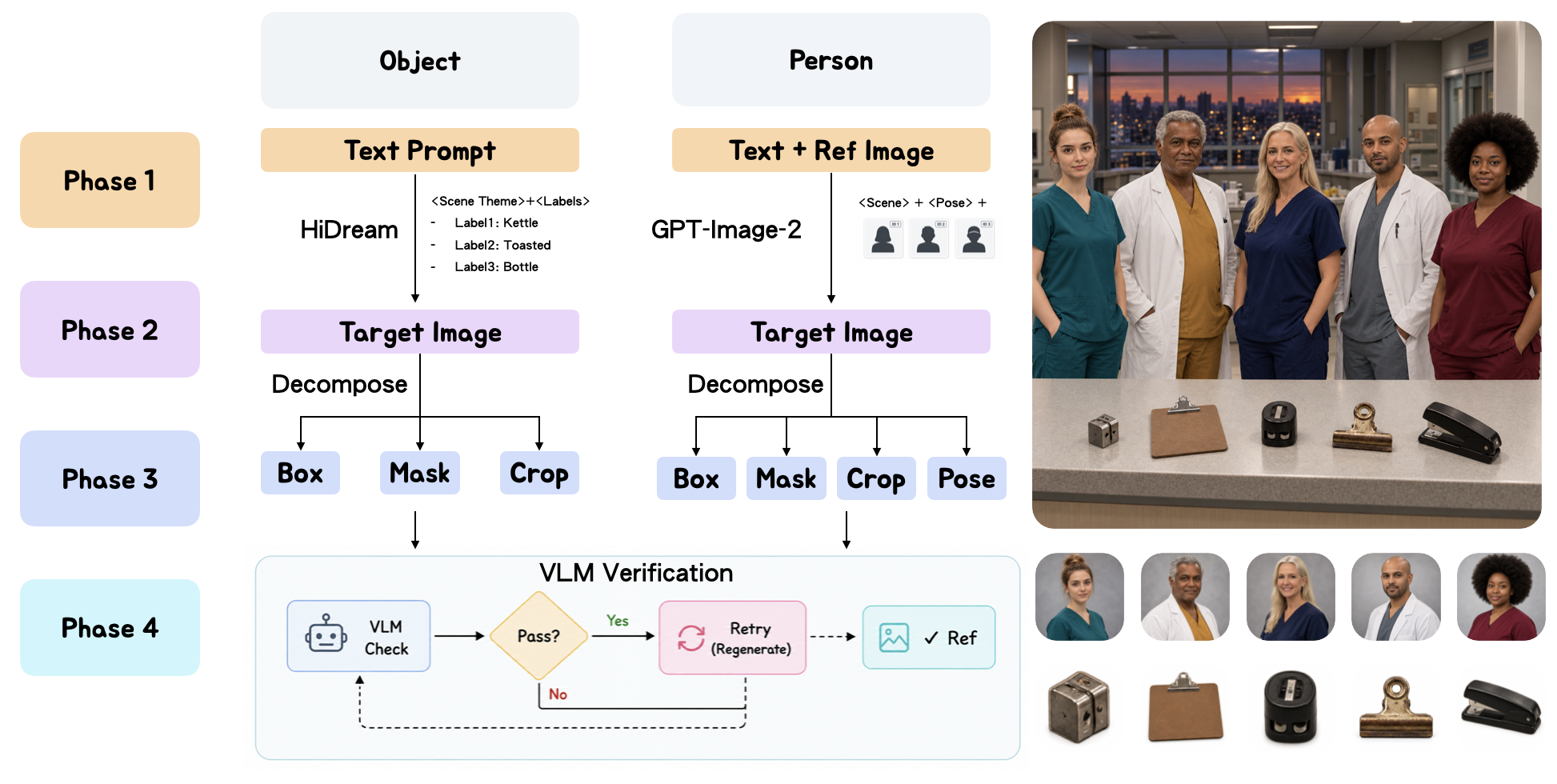}
\caption{\textbf{Schematic of the object and person construction branches.}
The diagram shows target synthesis, decomposition into boxes, masks, and
references (plus person poses), followed by verification. The illustrated
images explain the flow; corpus counts and quality limits are specified in
Appendix~\ref{sec:appendix:data-construction}.}
\label{fig:app-pipeline}
\end{figure}

Figure~\ref{fig:app-pipeline} summarizes the two construction paths.
In the audited examples, an object scene retains all eight requested
instances; a four-person identity-conditioned scene retains four matched
original references; and an eight-person text-conditioned scene exports
seven generated references, one flagged as identity-uncertain. These cases
illustrate the distinction between spatial annotations, reference
availability, and reference quality; they do not measure corpus-level
annotation accuracy.

\begingroup
\subsection{Evaluation Setup and Baselines}
\label{sec:appendix:max-reproducibility}

\paragraph{Models and comparisons.}
Table~\ref{tab:max-inference} summarizes the models used in our evaluation
and the source of each comparison.

\begin{table}[t]
\centering
\caption{\textbf{Models and baseline configurations.}}
\label{tab:max-inference}
\small
\setlength{\tabcolsep}{3pt}
\renewcommand{\arraystretch}{1.2}
\begin{tabular}{@{}>{\raggedright\arraybackslash}p{0.25\linewidth}>{\raggedright\arraybackslash}p{0.70\linewidth}@{}}
\toprule
Group & Models and use \\
\midrule
Ours & \texttt{Ours-MICo} (40k); \texttt{Ours-ManyRef} (high-$K$
continuation of \texttt{Ours-MICo}). Evaluated by us on MICo-Bench
and \ourbm{} at $1024^2$. \\
MICo-tuned open models & reported
MICo-Bench scores. \\
Native open models & FLUX.2-Klein-9B (evaluated by us); OmniGen2 and
Qwen-Image-Edit (reported MICo-Bench scores). \\
OminiControl2 mechanism & Released reference-branch code reproduced on our
backbone; no updated public weights. 4- and 50-step cost only. \\
\bottomrule
\end{tabular}
\end{table}
Published MICo-Bench scores follow each source's inference settings; they
are benchmark comparisons, not controlled inference ablations.

\paragraph{FLUX.2-Klein checkpoints.}
Throughout the paper, FLUX.2-Klein-9B refers to the FLUX.2 [klein] 9B
model family, and the number of sampling steps identifies the checkpoint.
Four-step results use the step-distilled FLUX.2-Klein-9B checkpoint, and
50-step results use the undistilled FLUX.2-Klein-Base-9B checkpoint.
Both checkpoints share the same architecture and parameter count. This
applies to Figure~\ref{fig:teaser}, Figure~\ref{fig:scaling}, and all
latency and speedup comparisons.

\paragraph{Quality evaluation.}
For our MICo-Bench ($N=897$) and \ourbm{} ($N=100$) evaluations, Overall
Weighted-Ref-VIEScore averages the per-case products:
\begin{equation}
Q=\frac{1}{N}\sum_{i=1}^{N}W_i\,SC_i\,PQ_i.
\label{eq:shared-quality-score}
\end{equation}
The MICo-Bench table also includes externally reported scores, labeled
separately in Table~\ref{tab:mico}. DINO, ArcFace, mIoU, and RRBA are
reported only for the residual-design ablation in Table~\ref{tab:residual}.

\paragraph{Inference cost.}
The reference-count curves use $K=1,2,4,8,12,16$ with reference KV caching
disabled. Latency is the median of three runs, and peak allocated memory
is the maximum over those runs. The OminiControl2 implementation reproduces
its reference-branch mechanism on the same backbone with an untrained
condition adapter; its measurements compare mechanism cost rather than
quality or the full trained system.
\par
\endgroup

\subsection{Effect of Stage~1 Pretraining}
\label{sec:appendix:stage1}
\begin{figure}[t]
\centering
\includegraphics[width=\linewidth]{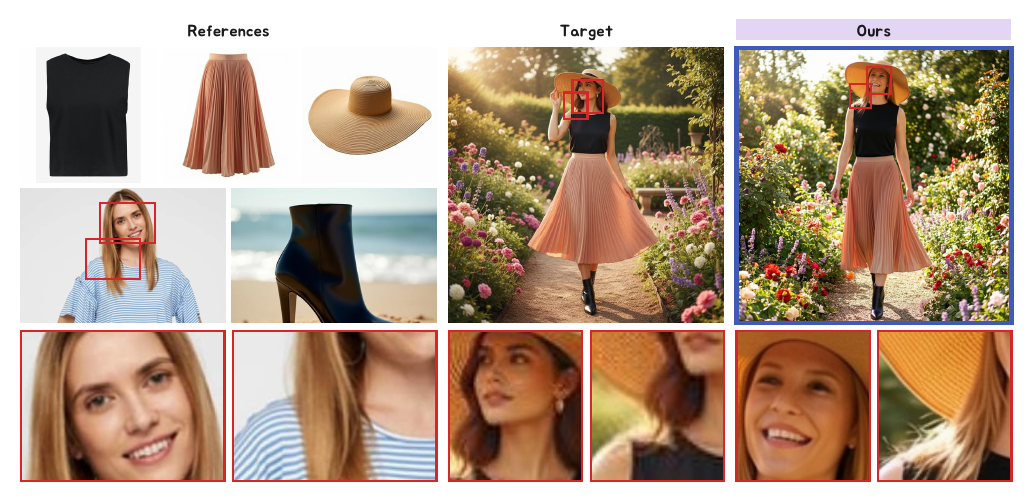}
\vspace{-0.2in}
\caption{\textbf{Stage~1 output.} Each reference is encoded as a
$\CompactGridSide\times\CompactGridSide$ grid plus a pixel residual.
After Stage~1, the model already preserves facial identity and garment
details. Target is the ground-truth image. Bottom: zoomed crops of the
red boxes.}
\vspace{-0.1in}
\label{fig:stage1}
\end{figure}

Figure~\ref{fig:stage1} shows what the compressor has learned by the end of Stage~1. From compact references, the model reproduces the facial identity of the person reference as well as the skirt pleats, hat brim, and boot shape. Even in the ground-truth target, the face and some details differ from the references, and the training signal is further averaged across references. Stage~1 training preserves these reference details faithfully.

\subsection{Pixel Residual Compressor Details}
\label{sec:appendix:compressor}
The pixel residual encoder uses \CompressorLayers\
$\CompressorKernel\times\CompressorKernel$ stride-$\CompressorStride$
convolutions with GroupNorm and SiLU activations, for a total downsampling
factor of $\CompressorStride^{\CompressorLayers}$.
Its output is aligned with the low-resolution reference grid, flattened, and
processed by \CompressorAttentionBlocks\ lightweight spatial self-attention blocks. A zero-initialized
linear projection maps the resulting features into the DiT hidden space,
where they are added to the projected low-resolution latent features.

\subsection{Training Configuration}
\label{sec:appendix:training}

\begin{table}[t]
\centering
\caption{\textbf{Training configuration by stage.}
Stage~2 uses 40k steps for the main experiments.
The high-$K$ variant continues training on \ourdata{} for
20k additional steps (60k total).
Stage~3 follows TBSM~\citep{sun2026tbsm} for distillation.
}
\label{tab:stages}
\footnotesize
\setlength{\tabcolsep}{3pt}
\renewcommand{\arraystretch}{1.12}
\begin{tabular}{@{}
p{0.18\linewidth}
p{0.23\linewidth}
p{0.31\linewidth}
p{\dimexpr0.28\linewidth-18pt\relax}
@{}}
\toprule
 & Stage 1 (T2I)
 & Stage 2 (joint training)
 & Stage 3 (distillation) \\
\midrule
Initialized from
 & Backbone weights
 & Stage 1 compressor + context LoRA
 & Trained teacher \\
Objective
 & $\mathcal{L}_{\mathrm{FM}}$ (text-to-image)
 & $\mathcal{L}_{\mathrm{FM}}$ (multi-ref, routed)
 & Distillation following TBSM~\citep{sun2026tbsm} \\
Trainable
 & Compressor, context LoRA
 & Compressor, context + spatial LoRA, layout stream
 & LoRA adapters \\
LoRA rank / LR
 & \LoRARank / $\TrainingLR$
 & \LoRARank / $\TrainingLR$
 & \LoRARank / $10^{-5}$ \\
Training steps
 & 40k
 & 40k (main); 60k total (high-$K$)
 & 20k \\
Batch / GPUs
 & $\PerGPUBatch{\times}\StageOneGPUs$ / \StageOneGPUs
 & $\PerGPUBatch{\times}\StageTwoGPUs$ / \StageTwoGPUs
 & $\PerGPUBatch{\times}\StageTwoGPUs$ / \StageTwoGPUs \\
Resolution
 & 1024 / 1024 
 & 1024 / 1024 
 & 1024 / 1024  \\
Reference count $K$
 & 1 
 & 1-5
 & > 5 \\
\bottomrule
\end{tabular}
\end{table}

\paragraph{Four-step inference.}
The recorded EMA36k student is measured at four sampling steps with CFG~1
and reference KV caching disabled. At $K=16$, its total latency is 7.84\,s,
compared with 111.10\,s for the four-step FLUX.2-Klein-9B baseline.
This is a cost measurement; it does not establish the student's generation
quality or the latency of the 60k high-$K$ run.

\paragraph{OminiControl2 mechanism comparison.}
We implement the OminiControl2 reference-branch mechanism~\citep{tan2025ominicontrol2}
on the same backbone, using native-resolution references
(4096 tokens per reference). Our representation uses 256 tokens per reference
with pixel residuals. The OminiControl2 condition adapter is untrained;
compact encoding and cross-step reuse are disabled.
The results characterize mechanism cost rather than the complete trained
system. Reference-branch topology also differs, so the timing gap is not
attributed solely to token count.

\section{Additional Qualitative Results}
\label{sec:additional-qualitative}

\subsection{Results on MICo-Bench}
\begin{figure}[p]
\centering
\includegraphics[width=\linewidth]{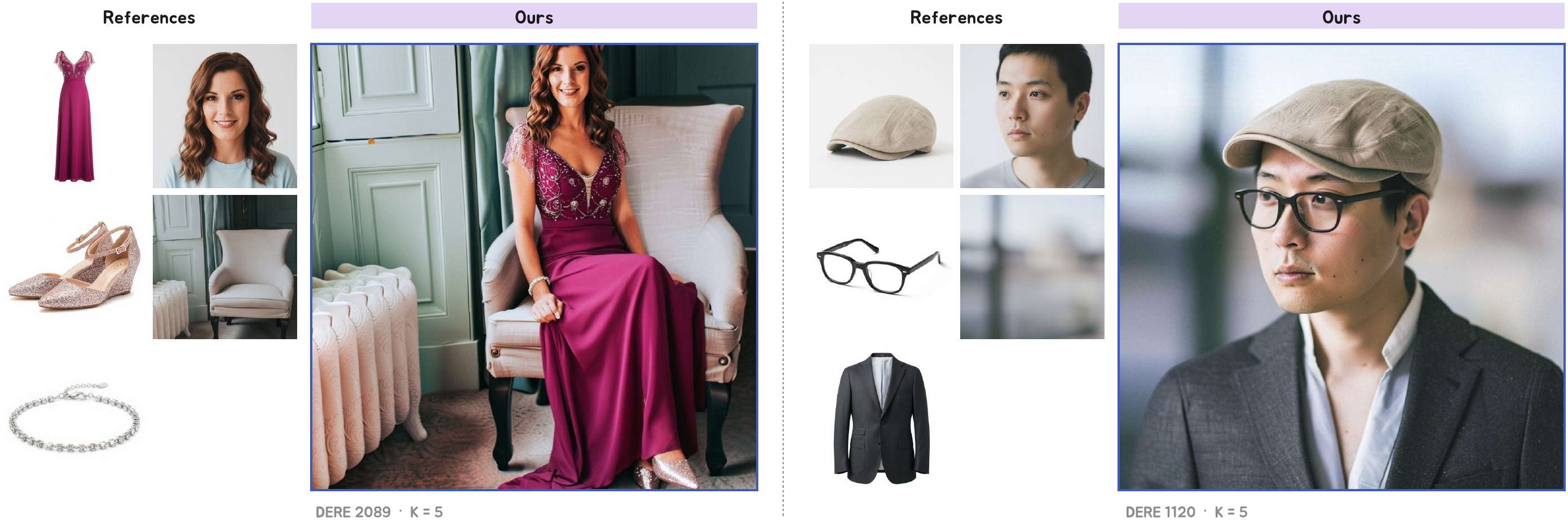}\\[2pt]
\includegraphics[width=\linewidth]{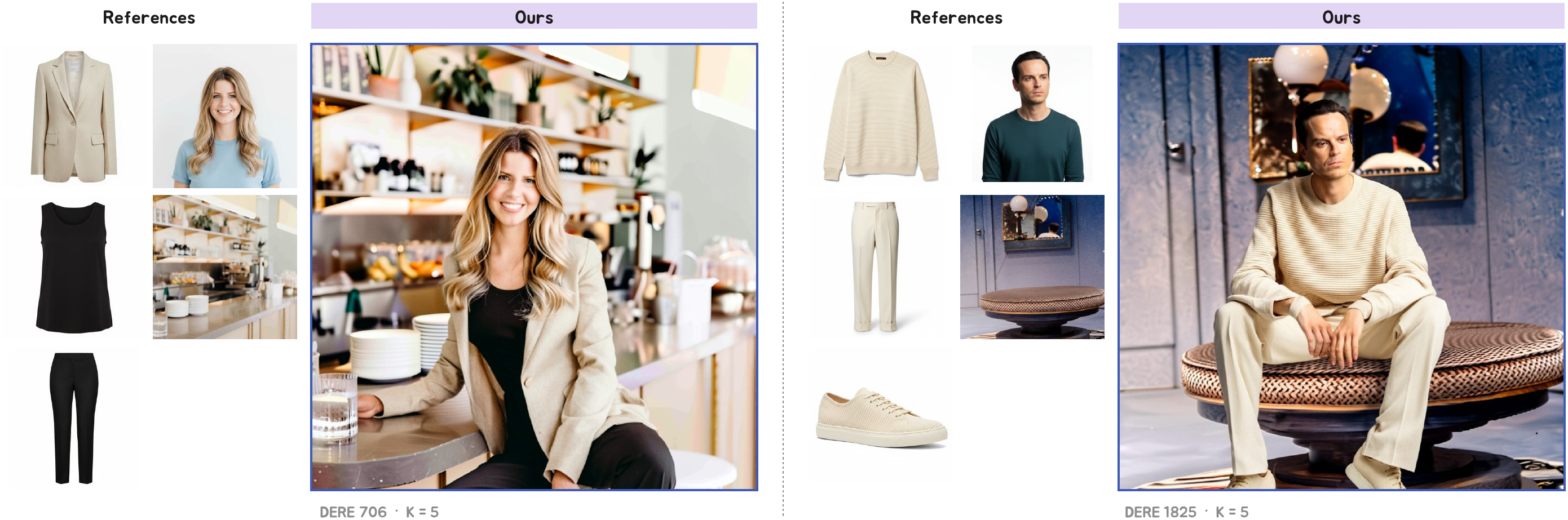}\\[2pt]
\includegraphics[width=\linewidth]{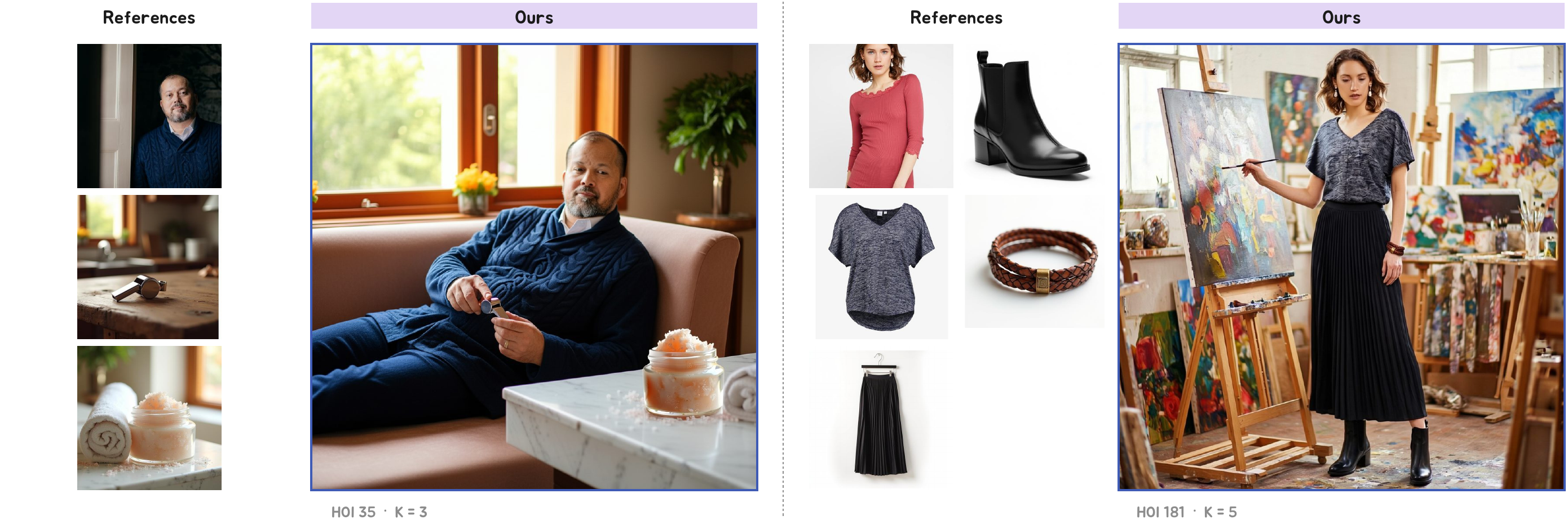}\\[2pt]
\includegraphics[width=\linewidth]{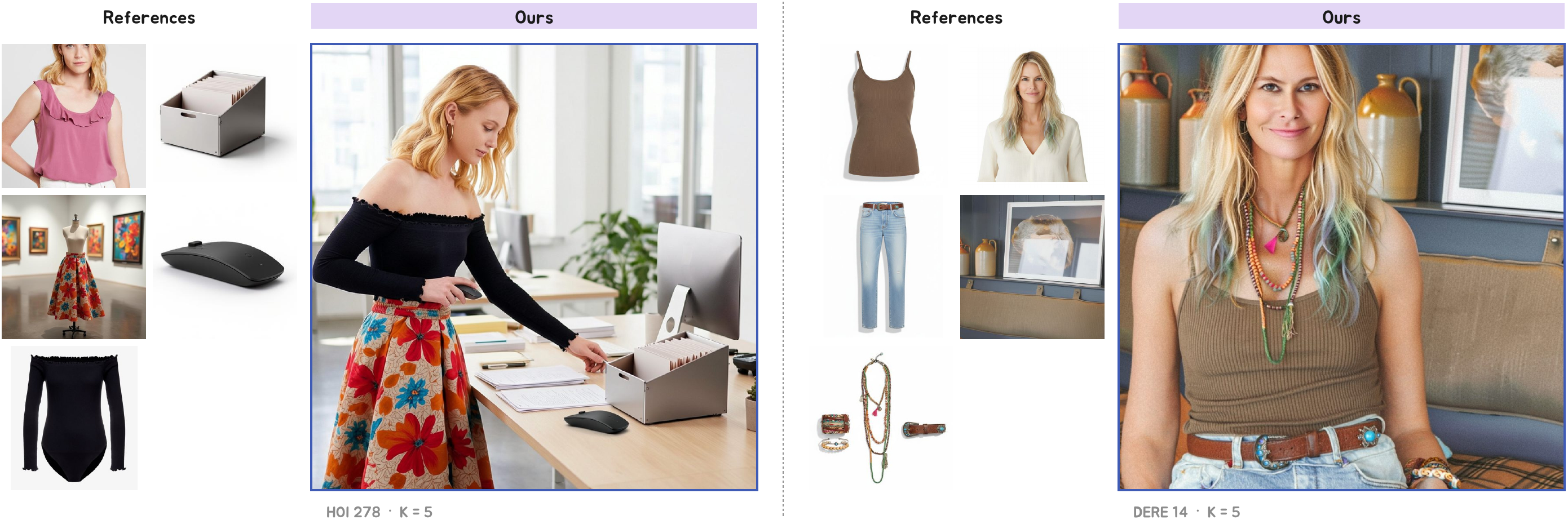}
\caption{Additional qualitative results on MICo-Bench.}
\label{fig:app-mico}
\end{figure}

Figure~\ref{fig:app-mico} shows eight additional MICo-Bench cases. Each
case shows the reference images (left) and our output (right); $K$ is the
number of references.

\FloatBarrier

\subsection{Additional Many-Reference Examples}
\begin{figure}[h]
\centering
\includegraphics[width=\linewidth]{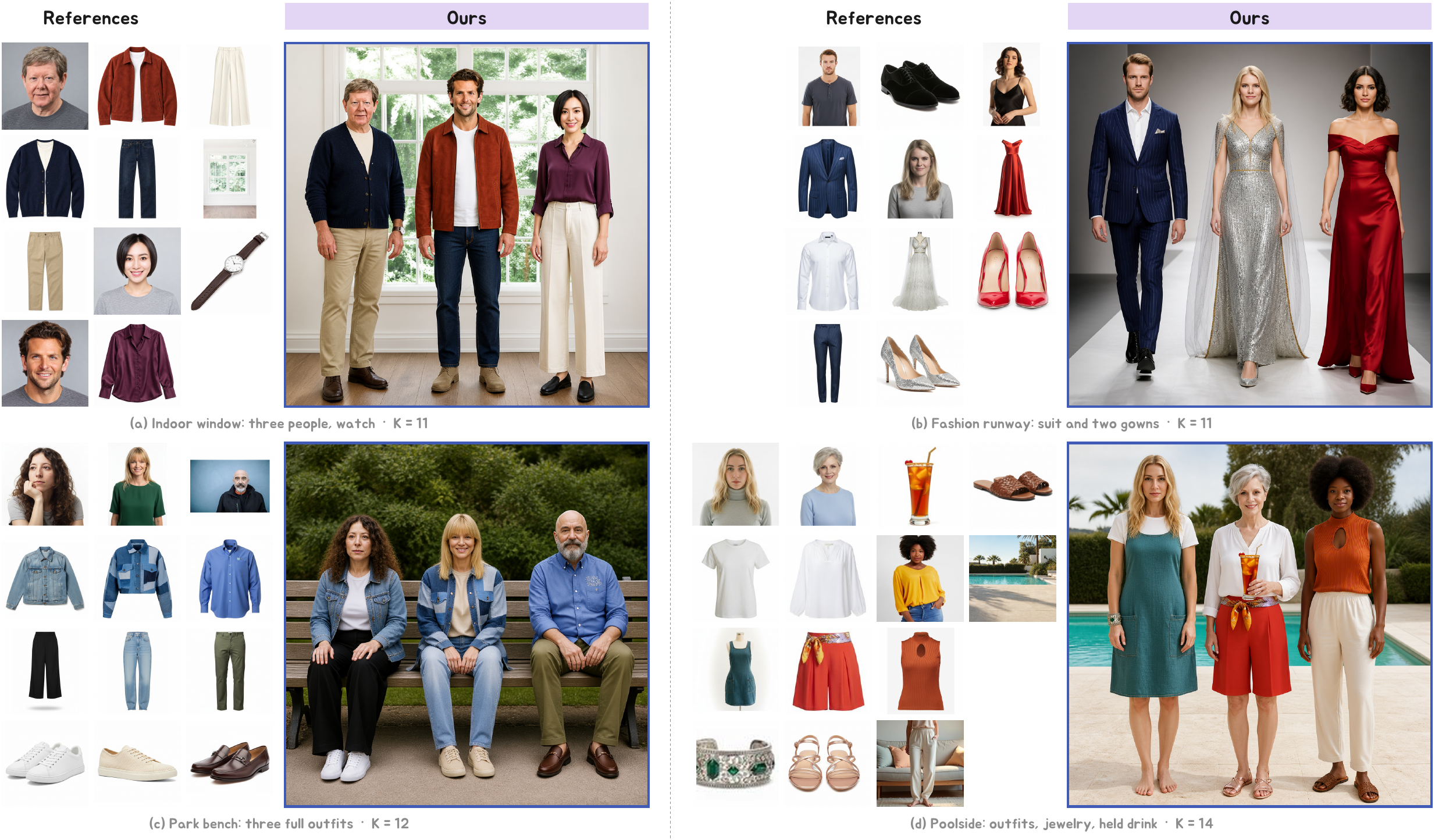}
\caption{Additional many-reference results with 11 to 14 references.}
\label{fig:app-manyref}
\end{figure}

Figure~\ref{fig:app-manyref} shows four additional many-reference
compositions with 11 to 14 references. Each case shows the reference
images (left) and our output (right).

\FloatBarrier

\end{document}